\let\accentvec\vec
\documentclass[runningheads,a4paper]{llncs}

\let\vec\accentvec

\usepackage{amsmath,amssymb,amsfonts,graphicx,url,marvosym}
\newcommand{\comment}[1]{}

\urldef{\mailsa}\path|{hy_zh, zlin}@pku.edu.cn|
\urldef{\mailsb}\path|chzhang@cis.pku.edu.cn|

\begin{document}
\mainmatter

\title{A Counterexample for the Validity of Using Nuclear Norm as
a Convex Surrogate of Rank}
\author{Hongyang Zhang\and Zhouchen Lin\thanks{Zhouchen Lin is the corresponding author.}\and Chao Zhang}
\institute{Key Lab. of Machine Perception (MOE), School of EECS\\
Peking University, Beijing, China\\
\mailsa, \mailsb}
\titlerunning{A Counterexample for the Validity of a Convex Surrogate}
\authorrunning{H. Zhang et al.}
\maketitle

\begin{abstract}
Rank minimization has attracted a lot of attention due to its robustness in data recovery. To overcome the computational difficulty, rank is often replaced with nuclear norm. For several rank minimization problems, such a replacement has been theoretically proven to be valid, i.e., the solution to nuclear norm minimization problem is also the solution to rank minimization problem. Although it is easy to \emph{believe} that such a replacement may not always be valid, no concrete example has ever been found. We argue that such a validity checking cannot be done by numerical computation and show, by analyzing the noiseless latent low rank representation (LatLRR) model, that even for very simple rank minimization problems the validity may still break down. As a by-product, we find that the solution to the nuclear norm minimization formulation of LatLRR is \emph{non-unique}. Hence the results of LatLRR reported in the literature may be questionable.
\end{abstract}

\section{Introduction}
\label{Section: Introduction}

We are now in an era of big data as well as high dimensional data.
Fortunately, high dimensional data are not unstructured. Usually,
they lie near low dimensional manifolds. This is the basis of
linear and nonlinear dimensionality
reduction~\cite{Wang:Geometry}. As a simple yet effective
approximation, linear subspaces are usually adopted to model the
data distribution. Because low dimensional subspaces correspond to
low rank data matrices, rank minimization problem, which models the real problem into an optimization by minimizing the rank in the objective function (cf. models \eqref{equation: noiseless rank LRR}, \eqref{equation: unobserved LatLRR} and \eqref{equation: noiseless original LatLRR}), is now widely used in
machine learning and data
recovery~\cite{Candes2,Candes,Candes3,Fazel}. Actually, rank is
regarded as a sparsity measure for matrices~\cite{Candes}. So low
rank recovery problems are
studied~\cite{DG:MC6,Wright:Compressive,Waters:SpaRCS,Liu:Universal}
in parallel with the compressed sensing theories for sparse vector
recovery. Typical rank minimization problems include matrix
completion~\cite{Candes2,Candes3}, which aims at completing the entire matrix from a small sample of its entries, robust principal component
analysis~\cite{Candes}, which recovers the ground truth data from sparsely corrupted elements, and low rank
representation~\cite{LiuG1,LiuG2}, which finds an affinity matrix of subspaces that has the lowest rank. All of these techniques have found wide
applications, such as background modeling~\cite{Candes}, image
repairing~\cite{Peng}, image alignment~\cite{Peng}, image
rectification~\cite{ZhangZ1}, motion
segmentation~\cite{LiuG1,LiuG2}, image
segmentation~\cite{Cheng:Multi}, and saliency
detection~\cite{Lang:Saliency}.

Since the rank of a matrix is discrete, rank minimization problems are usually hard to solve. They can even be NP
hard~\cite{Candes}. To overcome the computational obstacle, as a
common practice people usually replace rank in the objective function with nuclear norm,
which is the sum of singular values and is the convex envelope of
rank on the unit ball of matrix operator norm~\cite{Fazel}, to
transform rank minimization problems into nuclear norm minimization problems (cf. models \eqref{equation: noiseless nuclear LRR} and \eqref{equation: noiseless nuclear LatLRR}).
Such a strategy is widely adopted in most rank minimization
problems~\cite{Candes2,Candes,Candes3,LiuG1,LiuG2,Peng,ZhangZ1,Cheng:Multi,Lang:Saliency}.
However, this naturally brings a \emph{replacement validity problem} which is defined as follows.
\begin{definition}[Replacement Validity Problem]
Given a rank minimization problem together with its corresponding nuclear norm formulation, the replacement validity problem investigates whether the solution to the nuclear norm minimization problem is also a solution to the rank minimization one.
\end{definition}
In this paper, we focus on the replacement validity problem. There is a related problem, called \emph{exact recovery problem}, that is more widely studied by scholars. It is defined as follows.
\begin{definition}[Exact Recovery Problem]
Given a nuclear norm minimization problem, the exact recovery problem investigates the sufficient conditions under which the nuclear norm minimization problem could exactly recover the real structure of the data.
\end{definition}
As an example of the exact recovery problem, Cand\`{e}s et al. proved that when the rank of optimal solution is sufficiently low
and the missing data is sufficiently few or the corruption is
sufficiently sparse, solving nuclear norm minimization problems of matrix
completion~\cite{Candes2} or robust PCA problems~\cite{Candes} can
exactly recover the ground truth low rank solution with an
overwhelming probability. As another example, Liu et al.~\cite{LiuG1,Liu:Exact} proved
that when the rank of optimal solution is sufficiently low and the
percentage of corruption does not exceed a threshold, solving the
nuclear norm minimization problem of low rank representation (LRR)~\cite{LiuG1,LiuG2} can
exactly recover the ground truth subspaces of the data.

We want to highlight the difference between our replacement validity
problem and the exact recovery problem that scholars
have considered before. The replacement validity
problem is to compare the solutions
between two optimization problems, while the exact recovery problem is to study whether
solving a nuclear norm minimization problem can exactly recover a ground truth low rank
matrix. As a result, in all the existing exact recovery problems, the scholars have
to assume that the rank of the ground truth solution is sufficiently
low. In contrast, the replacement validity
problem does not rely on this assumption: even
if the ground truth low rank solution cannot be recovered, we can
still investigate whether the solution to a nuclear norm minimization problem is also the
solution to the corresponding rank minimization problem.

For replacement validity
problems, it is easy to \emph{believe} that the
replacement of rank with nuclear norm will break down for complex rank minimization problems. While for
exact recovery problems, the existing analysis all focuses on relatively simple rank minimization problems,
such as matrix completion~\cite{Candes2}, robust PCA
problems~\cite{Candes}, and LRR~\cite{LiuG1,LiuG2}, and has
achieved affirmative results under some conditions. So it is also
easy to \emph{believe} that for simple rank minimization problems the replacement of
rank with nuclear norm will work. This paper aims at breaking such
an illusion. Here, we have to point out that replacement validity
problem \emph{cannot} be studied by numerical experiments. This is because: 1. rank is
sensitive to numerical errors. Without prior knowledge, one may
not correctly determine the rank of a given matrix, even if there
is a clear drop in its singular values; 2. it is hard to verify
whether a given solution to nuclear norm minimization problem is a \emph{global} minimizer to
a rank minimization problem, whose objective function is discrete and non-convex. So we
should study replacement validity
problem by purely theoretical analysis. We analyze a
simple rank minimization problem, noiseless latent LRR (LatLRR)~\cite{LiuG3}, to show
that solutions to a nuclear norm minimization problem may not be solutions of the
corresponding rank minimization problem.

The contributions of this paper include:
\begin{enumerate}
\item We use a simple rank minimization problem, noiseless LatLRR, to prove that
solutions to a nuclear norm minimization problem may not be solutions of the corresponding
rank minimization problem, even for very simple rank minimization problems. \item As a by-product, we find
that LatLRR is not a good mathematical model because the solution
to its nuclear norm minimization formulation is \emph{non-unique}. So the results of
LatLRR reported in the literature, e.g., \cite{LiuG3,LiuG1}, may
be questionable.
\end{enumerate}

\section{Latent Low Rank Representation}
\label{section: Previous Works} \comment{ In this section, we
review a number of existing methods for related low rank problems.
Just as the discussion in Section \ref{Section: Introduction},
some of these models could be given complete closed form solutions
at least in the noiseless case.

In the noisy situation, for precisely recovering the intrinsic
structure, most of the methods add a regularization term to resist
noises, thus leading to higher model complexity in practice. In
this paper, we make a relaxed assumption on the data distributions
that they are completely noise-free. By removing the
regularization term, theoretical analysis are made possible for
deeper learning the structure of some low rank problems in the
meaning of their solutions.}

In this section, we first explain the notations that will be used
in this paper and then introduce latent low rank representation
which we will analyze its closed form solutions.

\subsection{Summary of Main Notations}
A large amount of matrix related symbols will be used in this
paper. Capital letters are used to represent matrices. Especially,
$I$ denotes the identity matrix and $0$ is the all-zero matrix. The
entry at the $i$th row and the $j$th column of a matrix is denoted
by $[\cdot]_{ij}$.  Nuclear norm, the sum of all the singular
values of a matrix, is denoted by $||\cdot||_*$. Operator norm,
the maximum singular value, is denoted by $||\cdot||_2$.
Trace($A$) represents the sum of the diagonal entries of $A$ and
$A^\dag$ is the Moore-Penrose pseudo-inverse of $A$. For
simplicity, we use the same letter to present the subspace spanned
by the columns of a matrix. The dimension of a space $V$ is
presented by $\mbox{dim}(V)$. The orthogonal complement of $V$ is
denoted by $V_\perp$. Range$(A)$ indicates the linear space
spanned by all the columns of matrix $A$, while Null($A$)
represents the null space of $A$. They are closely related:
$(\mbox{Range}(A))_\perp=\mbox{Null}(A^T)$. Finally, we always use
$U_X\Sigma_X V_X^T$ to represent the \emph{skinny} SVD of the data
matrix $X$. Namely, the numbers of columns in $U_X$ and $V_X$ are
both $\mbox{rank}(X)$ and $\Sigma_X$ consists of all the non-zero
singular values of $X$, making $\Sigma_X$ invertible.

\subsection{Low Rank Subspace Clustering Models}
Low rankness based subspace clustering stems from low rank
representation (LRR)~\cite{LiuG1,LiuG2}. An interested reader may refer to an excellent review on subspace clustering approaches provided by Vidal \cite{Vidal}. The mathematical model of
the original LRR is
\begin{equation}
\label{equation: noiseless rank LRR}
\min_Z \mbox{rank}(Z),\ \ \mbox{s.t.}\ \ X=XZ,
\end{equation}
where $X$ is the data matrix we observe. LRR extends sparse subspace clustering~\cite{Elhamifar} by
generalizing the sparsity from 1D to 2D. When there is noise or
corruption, a noise term can be added to the
model~\cite{LiuG1,LiuG2}. Since this paper considers closed form
solutions for noiseless models, to save space we omit the noisy
model. The corresponding nuclear norm minimization formulation of \eqref{equation:
noiseless rank LRR} is
\begin{equation}
\label{equation: noiseless nuclear LRR} \min_Z ||Z||_*,\ \
\mbox{s.t.}\ \ X=XZ,
\end{equation}
which we call the heuristic LRR. LRR has been very successful in
clustering data into subspaces
robustly~\cite{Adler:Probabilistic}. It is proven that when the
underlying subspaces are independent, the optimal representation
matrix is block diagonal, each block corresponding to a
subspace~\cite{LiuG1,LiuG2}.

\comment{

There have existed a large amount of works discussing the solution
structure of problem \eqref{equation: noiseless nuclear LRR}. In
\cite{LiuG2}, the authors demonstrated its block diagonal solution
if the data points strictly lie in the independent subspaces. More
generally, \cite{Lu} gave sufficient conditions for block diagonal
form of the optimal $Z$. \cite{LiuG1} also made a specific
conclusion: The solution to problem \eqref{equation: noiseless
nuclear LRR} is unique and takes the form $X^\dag X$, which is
called Shape Interaction Matrix (SIM). Besides that, the authors
also pointed out, as the sampling rate is not sufficient, SIM
degenerates to the trivial solution $I$, thus making LRR invalid.
Although so many researches have been done on the heuristic version
\eqref{equation: noiseless nuclear LRR}, few studies are conducted
on the problem \eqref{equation: noiseless rank LRR}, and as a
result remain a blank for complete theoretical analysis. In this
paper, we discuss complete solutions to original LRR via matrix
decomposition skill. As the most special case, we show that SIM is
also one of the optimal solutions to problem \eqref{equation:
noiseless rank LRR}, hence indirectly illustrating the relevance
between the rank function and the nuclear norm at least for LRR
model.

}

LRR works well only when the samples are sufficient. This
condition may not be fulfilled in practice, particularly when the
dimension of samples is large. To resolve this issue, Liu et
al.~\cite{LiuG3} proposed latent low rank representation (LatLRR).
Another model to overcome this drawback of LRR is fixed rank
representation~\cite{Liu:Fixed}. LatLRR assumes that the observed
samples can be expressed as the linear combinations of themselves together with the unobserved
data:
\begin{equation}
\label{equation: unobserved LatLRR}
\min_Z \mbox{rank}(Z),\ \ \mbox{s.t.}\ \ X=[X,X_H]Z,
\end{equation}
where $X_H$ is the unobserved samples for supplementing the
shortage of the observed ones. Since $X_H$ is unobserved and problem \eqref{equation: unobserved LatLRR} cannot be solved directly, by some
deduction and mathematical approximation, LatLRR~\cite{LiuG3} is
modeled as follows:
\begin{equation}
\label{equation: noiseless original LatLRR} \min_{Z,L}
\mbox{rank}(Z)+\mbox{rank}(L),\ \ \mbox{s.t.}\ \ X=XZ+LX.
\end{equation}
Both the optimal $Z$ and $L$ can be utilized for learning tasks:
$Z$ can be used for subspace clustering, while $L$ is for feature
extraction, thus providing us with the possibility for integrating
two tasks into a unified framework. We call \eqref{equation:
noiseless original LatLRR} the original LatLRR. Similarly, it has
a nuclear norm minimization formulation
\begin{equation}
\label{equation: noiseless nuclear LatLRR} \min_{Z,L}
||Z||_*+||L||_*,\ \ \mbox{s.t.}\ \ X=XZ+LX,
\end{equation}
which we call the heuristic LatLRR. LatLRR has been reported to have
better performance than LRR~\cite{LiuG3,LiuG1}.

In this paper, we focus on studying the solutions to problems \eqref{equation: noiseless rank LRR}, \eqref{equation: noiseless nuclear LRR}, \eqref{equation: noiseless original LatLRR} and \eqref{equation: noiseless nuclear LatLRR}, in order to investigate the replacement validity problem.

\comment{

Formalized as a convex optimization, problem \eqref{equation:
noiseless nuclear LatLRR} is often globally solved by the method
of augmented Lagrange multipliers (ALM). Unlike LRR model, which
fails when the sampling is insufficient, the numerical experiments
conducted by ALM show a block diagonal solution to LatLRR
\cite{LiuG3}, and as a result verify the validity of model
\eqref{equation: noiseless nuclear LatLRR}.

Although so much success has been realized in the real
applications, to the best of our knowledge, no theoretical
analysis has been carried out on the problem \eqref{equation:
noiseless original LatLRR} and \eqref{equation: noiseless nuclear
LatLRR}, thus having not illustrated why LatLRR is so valid even
when the sampling is not enough. In this paper, we give the
sufficient and necessary conditions for the closed form solutions
to \eqref{equation: noiseless original LatLRR} and
\eqref{equation: noiseless nuclear LatLRR}. Unlike ALM numerical
method, which searches for only one optimality, our theory has
nothing to do with this limit and successfully provides the entire
solutions to LatLRR model with deeper insight.

}

\section{Analysis on LatLRR}
\label{section: Analysis on Rank Minimization Model} This
section provides surprising results: both the original and
heuristic LatLRR have closed form solutions! We are able to write
down \emph{all} their solutions, as presented in the following
theorems.

\comment{So we can determine whether the solutions to the nuclear norm minimization
formulation of LatLRR are also the solutions to the rank minimization problem one. We
also find that the solution to the heuristic LatLRR is
\emph{non-unique}. So the results of LatLRR reported in the
literature, e.g., \cite{LiuG3,LiuG2}, may be questionable.

The \emph{complete} closed form solutions to the original and
heuristic LatLRR are given by the following theorems, respectively.
}

\begin{theorem}
\label{theorem: complete solution to noiseless rank LatLRR} The
complete solutions to the original LatLRR problem \eqref{equation:
noiseless original LatLRR} are as follows
\begin{equation}
\label{equation: solution to noiseless original LatLRR}
Z^*=V_X\tilde{W}V_X^T+S_1\tilde{W}V_X^T  \mbox{ and }
L^*=U_X\Sigma_X(I-\tilde{W})\Sigma_X^{-1}U_X^T+U_X\Sigma_X(I-\tilde{W})S_2,
\end{equation}
where $\tilde{W}$ is any idempotent matrix and $S_1$ and $S_2$ are
any matrices satisfying: 1. $V_X^TS_1=0$ and $S_2U_X=0$; and 2.
$\mbox{rank}(S_1)\leq \mbox{rank}(\tilde{W})$ and
$\mbox{rank}(S_2)\leq \mbox{rank}(I-\tilde{W})$.
\end{theorem}

\begin{theorem}
\label{theorem: sufficient and necessary condition for nuclear
norm LatLRR} The complete solutions to the heuristic LatLRR problem
\eqref{equation: noiseless nuclear LatLRR} are as follows
\begin{equation}
\label{equation: solution to noiseless heuristic LatLRR}
Z^*=V_X\widehat{W}V_X^T \mbox{ and } L^*=U_X(I-\widehat{W})U_X^T,
\end{equation}
where $\widehat{W}$ is any block diagonal matrix satisfying: 1. its
blocks are compatible with $\Sigma_X$, i.e., if
$[\Sigma_X]_{ii}\neq[\Sigma_X]_{jj}$ then $[\widehat{W}]_{ij}=0$; and
2. both $\widehat{W}$ and $I-\widehat{W}$ are positive semi-definite.
\end{theorem}

By Theorems~\ref{theorem: complete solution to noiseless rank
LatLRR} and~\ref{theorem: sufficient and necessary condition for
nuclear norm LatLRR}, we can conclude that if the $\widehat{W}$ in
Theorem~\ref{theorem: sufficient and necessary condition for
nuclear norm LatLRR} is not idempotent, then the corresponding
$(Z^*,L^*)$ is not the solution to the original LatLRR, due to the
following proposition:

\begin{proposition}
\label{proposition:vality} If the $\widehat{W}$ in
Theorem~\ref{theorem: sufficient and necessary condition for
nuclear norm LatLRR} is not idempotent, then $Z^*=V_X\widehat{W}V_X^T$
cannot be written as $Z^*=V_X\tilde{W}V_X^T+S_1\tilde{W}V_X^T$,
where $\tilde{W}$ and $S_1$ satisfy the conditions stated in
Theorem~\ref{theorem: complete solution to noiseless rank LatLRR}.
\end{proposition}

The above results show that for noiseless LatLRR, nuclear norm is
not a valid replacement of rank. As a by-product, since the
solution to the heuristic LatLRR is \emph{non-unique}, the results of
LatLRR reported in~\cite{LiuG3,LiuG2} may be questionable.

We provide detailed proofs of the above theorems and proposition
in the following section.

\section{Proofs}\label{section:proofs}
\subsection{Proof of Theorem \ref{theorem: complete solution to
noiseless rank LatLRR}}

We first provide the complete closed form solutions to the
original LRR in a more general form
\begin{equation}
\label{equation: noiseless generalized rank LRR}
\min_Z \mbox{rank}(Z),\ \ \mbox{s.t.}\ \ A=XZ,
\end{equation}
where $A\in\mbox{Range}(X)$ so that the constraint is feasible. We
call \eqref{equation: noiseless generalized rank LRR} the
generalized original LRR. Then we have the following proposition.

\begin{proposition}
\label{proposition: complete solution to generalized rank LRR}
Suppose $U_A\Sigma_AV_A^T$ is the skinny SVD of $A$. Then the
minimum objective function value of the generalized original LRR problem
\eqref{equation: noiseless generalized rank LRR} is
$\mbox{rank}(A)$ and the complete solutions to \eqref{equation:
noiseless generalized rank LRR} are as follows
\begin{equation}
\label{equation: optimal Z to noiseless rank LRR} Z^*=X^\dag
A+SV_A^T,
\end{equation}
where $S$ is any matrix such that $V_X^TS=0$.
\end{proposition}
\begin{proof}
Suppose $Z^*$ is an optimal solution to problem \eqref{equation:
noiseless generalized rank LRR}. First, we have
\begin{equation}
\label{equation: rank(A)<=rank(Z)}
\mbox{rank}(A)=\mbox{rank}(XZ^*)\le\mbox{rank}(Z^*).
\end{equation}
On the other hand, because $A=XZ$ is feasible, there exists $Z_1$
such that $A=XZ_1$. Then $Z_0=X^\dag A$ is feasible:
$XZ_0=XX^\dag A= XX^\dag XZ_1 = XZ_1=A$, where we have utilized a
property of Moore-Penrose pseudo-inverse $XX^\dag X=X$. So we
obtain
\begin{equation}
\label{equation: rank(Z)<=rank(A)}
\mbox{rank}(Z^*)\le\mbox{rank}(Z_0)\le\mbox{rank}(A).
\end{equation}
Combining \eqref{equation: rank(A)<=rank(Z)} with \eqref{equation:
rank(Z)<=rank(A)}, we conclude that $\mbox{rank}(A)$ is the
minimum objective function value of problem \eqref{equation:
noiseless generalized rank LRR}.

Next, let $Z^*=PQ^T$ be the full rank decomposition of the optimal
$Z^*$, where both $P$ and $Q$ have rank$(A)$ columns. From
$U_A\Sigma_AV_A^T=XPQ^T$, we have
$V_A^T=(\Sigma_A^{-1}U_A^TXP)Q^T$. Since both $V_A$ and $Q$ are
full column rank and $Y=\Sigma_A^{-1}U_A^TXP$ is square, $Y$ must
be invertible. So $V_A$ and $Q$ represent the same subspace.
Because $P$ and $Q$ are unique up to an invertible matrix, we may
simply choose $Q=V_A$. Thus $U_A\Sigma_AV_A^T=XPQ^T$ reduces to
$U_A\Sigma_A=U_X\Sigma_XV_X^TP$, i.e.,
$V_X^TP=\Sigma_X^{-1}U_X^TU_A\Sigma_A$, and we conclude that the
complete choices of $P$ are given by
$P=V_X\Sigma_X^{-1}U_X^TU_A\Sigma_A+S$, where $S$ is any matrix
such that $V_X^TS=0$. Multiplying $P$ with $Q^T=V_A^T$, we obtain
that the entire solutions to problem \eqref{equation: noiseless
generalized rank LRR} can be written as $Z^*=X^\dag A+SV_A^T$,
where $S$ is any matrix satisfying $V_X^TS=0$. $\Box$
\end{proof}

\begin{remark}
Friedland and Torokhti \cite{Friedland} studied a similar model as \eqref{equation: noiseless generalized rank LRR}, which is
\begin{equation}
\label{equation: similar problem as generalized rank LRR}
\min_Z ||X-AZ||_F,\ \ \mbox{s.t.}\ \ \mbox{rank}(Z)\le k.
\end{equation}
However, \eqref{equation: noiseless generalized rank LRR} is different from \eqref{equation: similar problem as generalized rank LRR} in two aspects. First, \eqref{equation: noiseless generalized rank LRR} requires the data matrix $X$ to be strictly expressed as linear combinations of the columns in $A$. Second, \eqref{equation: noiseless generalized rank LRR} does not impose an upper bound for the rank of $Z$. Rather, \eqref{equation: noiseless generalized rank LRR} solves for the $Z$ with the lowest rank. As a result, \eqref{equation: noiseless generalized rank LRR} has infinitely many solutions, as shown by Proposition \ref{proposition: complete solution to generalized rank LRR}, while \eqref{equation: similar problem as generalized rank LRR} has a unique solution when $k$ fulfills some conditions. So the results in \cite{Friedland} do not apply to \eqref{equation: noiseless generalized rank LRR}.
\end{remark}

Similar to Proposition \ref{proposition: complete solution to
generalized rank LRR}, we can have the complete closed form
solution to the following problem
\begin{equation}
\label{equation: noiseless generalized feature extraction} \min_Z
\mbox{rank}(L),\ \ \mbox{s.t.}\ \ A=LX,
\end{equation}
which will be used in the proof of Theorem \ref{theorem: complete
solution to noiseless rank LatLRR}.

\begin{proposition}
\label{proposition: complete solution to extending feature
extraction} Suppose $U_A\Sigma_AV_A^T$ is the skinny SVD of $A$.
Then the minimum objective function value of problem
\eqref{equation: noiseless generalized feature extraction} is
$\mbox{rank}(A)$ and the complete solutions to problem
\eqref{equation: noiseless generalized feature extraction} are as
follows
\begin{equation}
L^*=AX^\dag+U_AS,
\end{equation}
where $S$ is any matrix such that $SU_X=0$.
\end{proposition}

Next, we provide the following propositions.
\begin{proposition}
\label{proposition: the minimal objective to original LatLRR}
$\mbox{rank}(X)$ is the minimum objective function value of the
original LatLRR problem \eqref{equation: noiseless original
LatLRR}.
\end{proposition}
\begin{proof}
Suppose $(Z^*,L^*)$ is an optimal solution to problem
\eqref{equation: noiseless original LatLRR}. By
Proposition~\ref{proposition: complete solution to generalized
rank LRR} and fixing $Z^*$, we have
$\mbox{rank}(L^*)=\mbox{rank}(X-XZ^*)$. Thus
\begin{equation}
\mbox{rank}(Z^*)+\mbox{rank}(L^*)\ge\mbox{rank}(XZ^*)+\mbox{rank}(X-XZ^*)\ge\mbox{rank}(X).
\end{equation}
On the other hand, if $Z^*$ and $L^*$ are adopted as $X^\dag X$
and $0$, respectively, the lower bound is achieved and the
constraint is fulfilled as well. So we conclude that
$\mbox{rank}(X)$ is the minimum objective function value of the
original LatLRR problem \eqref{equation: noiseless original
LatLRR}. $\Box$
\end{proof}

\begin{proposition}
\label{proposition: XZ=XZ^*} Suppose $(Z^*,L^*)$ is one of the
solutions to problem \eqref{equation: noiseless original LatLRR}.
Then there must exist another solution
$(\widetilde{Z},\widetilde{L})$, such that $XZ^*=X\widetilde{Z}$
and $\widetilde{Z}=V_X\tilde{W}V_X^T$ for some matrix $\tilde{W}$.
\end{proposition}
\begin{proof}
According to the constraint of problem \eqref{equation: noiseless
original LatLRR}, we have $XZ=(I-L)X$, i.e.,
$(XZ)^T\in\mbox{Range}(X^T)$. Since $V_XV_X^T$ is the projection
matrix onto $\mbox{Range}(X^T)$, we have
\begin{equation}
\label{equation: projection} XZ^*V_XV_X^T=XZ^*.
\end{equation}

On the other hand, given the optimal $Z^*$, $L^*$ is the optimal
solution to
\begin{equation}
\label{equation: subproblem} \min_L \mbox{rank}(L)\ \ \mbox{s.t.}\
\ X(I-Z^*)=LX.
\end{equation}
So by Proposition \ref{proposition: complete solution to
generalized rank LRR} we get
\begin{equation}
\label{equation: rank equation on L}
\mbox{rank}(L^*)=\mbox{rank}(X(I-Z^*)X^\dag).
\end{equation}
As a result,
\begin{equation}
\label{equation: deduct on original LatLRR VWV^T}
\begin{split}
\mbox{rank}(X)&=\mbox{rank}(Z^*)+\mbox{rank}(L^*)\\
&=\mbox{rank}(Z^*)+\mbox{rank}(X(I-Z^*)X^\dag)\\
&=\mbox{rank}(Z^*)+\mbox{rank}(X(I-V_XV_X^TZ^*V_XV_X^T)X^\dag)\\
&\geq \mbox{rank}(V_XV_X^TZ^*V_XV_X^T)+\mbox{rank}(X(I-V_XV_X^TZ^*V_XV_X^T)X^\dag)\\
&\ge\mbox{rank}(X),
\end{split}
\end{equation}
where the last inequality holds since
$(V_XV_X^TZ^*V_XV_X^T,X(I-V_XV_X^TZ^*V_XV_X^T)X^\dag)$ is a
feasible solution to problem \eqref{equation: noiseless original
LatLRR} and $\mbox{rank}(X)$ is the minimum objective according to
Proposition~\ref{proposition: the minimal objective to original
LatLRR}. \eqref{equation: deduct on original LatLRR VWV^T} shows
that $(V_XV_X^TZ^*V_XV_X^T,X(I-V_XV_X^TZ^*V_XV_X^T)X^\dag)$ is an
optimal solution. So we may take
$\widetilde{Z}=V_XV_X^TZ^*V_XV_X^T$ and write it as
$\widetilde{Z}=V_X\tilde{W}V_X^T$, where $\tilde{W}=V_X^TZ^*V_X$.

Finally, combining with equation \eqref{equation: projection}, we
conclude that
\begin{equation}
X\widetilde{Z}=U_X\Sigma_X
V_X^TV_XV_X^TZ^*V_XV_X^T=XZ^*V_XV_X^T=XZ^*.
\end{equation} $\Box$
\end{proof}

Proposition \ref{proposition: XZ=XZ^*} provides us with a great
insight into the structure of problem \eqref{equation: noiseless
original LatLRR}: we may break \eqref{equation: noiseless original
LatLRR} into two subproblems
\begin{equation}
\label{equation: Z subproblem} \min_Z \mbox{rank}(Z),\ \
\mbox{s.t.}\ \ XV_X\tilde{W}V_X^T=XZ,
\end{equation}
and
\begin{equation}
\label{equation: L subproblem} \min_L \mbox{rank}(L),\ \
\mbox{s.t.}\ \ X-XV_X\tilde{W}V_X^T=LX,
\end{equation}
and then apply Propositions \ref{proposition: complete solution to
generalized rank LRR} and \ref{proposition: complete solution to
extending feature extraction} to find the complete solutions to
problem \eqref{equation: noiseless original LatLRR}.

For investigating the properties of $\tilde{W}$ in \eqref{equation:
Z subproblem} and \eqref{equation: L subproblem}, the following
lemma is critical.
\begin{lemma}
\label{lemma: rank relationship on exchanged A and B} For
$A,B\in\mathbb{R}^{n\times n}$, if $AB=BA$, then the following
inequality holds
\begin{equation}
\label{equation: inequality on rank and dimension of Null}
\mbox{rank}(A+B)\le\mbox{rank}(A)+\mbox{rank}(B)-\mbox{rank}(AB).
\end{equation}
\end{lemma}
\begin{proof}
On the basis of $AB=BA$, it is easy to check that
\begin{equation}
\label{equation: null(A)+null(B) and null(AB)}
\mbox{Null}(A)+\mbox{Null}(B)\subset\mbox{Null}(AB),
\end{equation}
and
\begin{equation}
\label{equation: null(A) cap null(B) and null(A+B)}
\mbox{Null}(A)\cap\mbox{Null}(B)\subset\mbox{Null}(A+B).
\end{equation}
On the other hand, according to the well-known dimension formula
\begin{equation}
\label{equation: dimensional formula}
\mbox{dim(Null}(A))+\mbox{dim(Null}(B))=\mbox{dim}(\mbox{Null}(A)+\mbox{Null}(B))+\mbox{dim}(\mbox{Null}(A)\cap\mbox{Null}(B)),
\end{equation}
by combining \eqref{equation: dimensional formula} with
\eqref{equation: null(A)+null(B) and null(AB)} and
\eqref{equation: null(A) cap null(B) and null(A+B)}, we get
\begin{equation}
\mbox{dim(Null}(A))+\mbox{dim(Null}(B))=\mbox{dim}(\mbox{Null}(AB))+\mbox{dim}(\mbox{Null}(A+B)).
\end{equation}
Then by the relationship $\mbox{rank}(S)=n-\mbox{dim(Null}(S))$
for any $S\in\mathbb{R}^{n\times n}$, we arrive at the inequality
\eqref{equation: inequality on rank and dimension of Null}. $\Box$
\end{proof}

Based on the above lemma, the following proposition presents the
sufficient and necessary condition on $\tilde{W}$.
\begin{proposition}
\label{proposition: perperty on W} Let $L^*$ be any optimal solution to subproblem \eqref{equation: L subproblem}, then
$(V_X\tilde{W}X_X^T,L^*)$ is optimal to problem \eqref{equation:
noiseless original LatLRR} if and only if the square matrix
$\tilde{W}$ is idempotent.
\end{proposition}
\begin{proof}
Obviously, $(V_X\tilde{W}X_X^T,L^*)$ is feasible based on the
constraint in problem \eqref{equation: L subproblem}. By
considering the optimality of $L^*$ for \eqref{equation: L
subproblem} and replacing $Z^*$ with $V_X\tilde{W}V_X^T$ in
equation \eqref{equation: rank equation on L}, we have
\begin{equation}
\mbox{rank}(L^*)=\mbox{rank}(X(I-V_X\tilde{W}V_X^T)X^\dag).
\end{equation}

First, we prove the sufficiency. According to the property of
idempotent matrices, we have
\begin{equation}
\mbox{rank}(\tilde{W})=\mbox{trace}(\tilde{W}) \mbox{ and }
\mbox{rank}(I-\tilde{W})=\mbox{trace}(I-\tilde{W}).
\end{equation}
By substituting $(V_X\tilde{W}V_X^T,L^*)$ into the objective
function, the following equalities hold
\begin{equation}
\begin{split}
\mbox{rank}(V_X\tilde{W}V_X^T)+\mbox{rank}(L^*)&=\mbox{rank}(\tilde{W})+\mbox{rank}(X(I-V_X\tilde{W}V_X^T)X^\dag)\\
&=\mbox{rank}(\tilde{W})+\mbox{rank}(U_X\Sigma_X(I-\tilde{W})\Sigma_X^{-1}U_X^T)\\
&=\mbox{rank}(\tilde{W})+\mbox{rank}(I-\tilde{W})\\
&=\mbox{trace}(\tilde{W})+\mbox{trace}(I-\tilde{W})\\
&=\mbox{rank}(X).
\end{split}
\end{equation}
So $(V_X\tilde{W}X_X^T,L^*)$ is optimal since it achieves the
minimum objective function value of problem \eqref{equation:
noiseless original LatLRR}.

Second, we prove the necessity. Suppose $(V_X\tilde{W}V_X^T,L^*)$
is optimal to problem \eqref{equation: noiseless original LatLRR}.
Substituting it into the objective follows
\begin{equation}
\begin{split}
\mbox{rank}(X)&=\mbox{rank}(V_X\tilde{W}X_X^T)+\mbox{rank}(X(I-V_X\tilde{W}V_X)X^\dag)\\
&=\mbox{rank}(\tilde{W})+\mbox{rank}(I-\tilde{W})\\
&\ge\mbox{rank}(X).
\end{split}
\end{equation}
Hence
$\mbox{rank}(\tilde{W})+\mbox{rank}(I-\tilde{W})=\mbox{rank}(X)$.
On the other hand, as $\tilde{W}$ and $I-\tilde{W}$ are
commutative, by Lemma \ref{lemma: rank relationship on exchanged A
and B} we have
$\mbox{rank}(X)\le\mbox{rank}(\tilde{W})+\mbox{rank}(I-\tilde{W})-\mbox{rank}(\tilde{W}-\tilde{W}^2)$.
So $\mbox{rank}(\tilde{W}-\tilde{W}^2)=0$ and thus
$\tilde{W}=\tilde{W}^2$. $\Box$
\end{proof}

We are now ready to prove Theorem \ref{theorem: complete solution
to noiseless rank LatLRR}.

\begin{proof}
Solving problems \eqref{equation: Z subproblem} and
\eqref{equation: L subproblem} by using Propositions
\ref{proposition: complete solution to generalized rank LRR} and
\ref{proposition: complete solution to extending feature
extraction}, where $\tilde{W}$ is idempotent as Proposition
\ref{proposition: perperty on W} shows, we directly get
\begin{equation}
\label{equation: simplest solution for rank LatLRR}
Z^*=V_X\tilde{W}V_X^T+\widetilde{S}_1V_A^T \mbox{ and } L^*=U_X\Sigma_X(I-\tilde{W})\Sigma_X^{-1}U_X^T+U_B\widetilde{S}_2,
\end{equation}
where $U_A\Sigma_AV_A^T$ and $U_B\Sigma_BV_B^T$ are the skinny
SVDs of $U_X\Sigma_X\tilde{W}V_X^T$ and
$U_X\Sigma_X(I-\tilde{W})V_X^T$, respectively, and $\widetilde{S}_1$
and $\widetilde{S}_2$ are matrices such that $V_X^T\widetilde{S}_1=0$ and
$\widetilde{S}_2U_X=0$. Since we have
$\mbox{Range}((\tilde{W}V_X^T)^T)=\mbox{Range}(V_A)$ and
$\mbox{Range}(U_X\Sigma_X(I-\tilde{W}))=\mbox{Range}(U_B)$, there
exist full column rank matrices $M_1$ and $M_2$ satisfying
$V_A=(\tilde{W}V_X^T)^TM_1$ and $U_B=U_X\Sigma_X(I-\tilde{W})M_2$,
respectively. The sizes of $M_1$ and $M_2$ are
$\mbox{rank}(X)\times\mbox{rank}(\tilde{W})$ and
$\mbox{rank}(X)\times\mbox{rank}(I-\tilde{W})$, respectively.
We can easily see that a matrix $S_1$ can be decomposed into $S_1=\widetilde{S}_1M_1^T$, such that $V_X^T\widetilde{S}_1=0$ and $ M_1$ is full column rank, if and only if $V_X^TS_1=0$ and $\mbox{rank}(S_1)\leq \mbox{rank}(\tilde{W})$. Similarly, a matrix $S_2$ can be decomposed into $S_2=M_2\widetilde{S}_2 $, such that $\widetilde{S}_2U_X=0$ and $ M_2$ is full column rank, if and only if $S_2U_X=0$ and $\mbox{rank}(S_2)\leq \mbox{rank}(I-\tilde{W})$. By substituting
$V_A=(\tilde{W}V_X^T)^TM_1$, $U_B=U_X\Sigma_X(I-\tilde{W})M_2$,
$S_1=\widetilde{S}_1M_1^T$, and $S_2=M_2\widetilde{S}_2$ into
\eqref{equation: simplest solution for rank LatLRR}, we obtain the
conclusion of Theorem \ref{theorem: complete solution to noiseless
rank LatLRR}. $\Box$
\end{proof}

\subsection{Proof of Theorem \ref{theorem: sufficient and necessary
condition for nuclear norm LatLRR}} We first quote two results
from~\cite{LiuG1}.

\begin{lemma}
\label{lemma: unique solution to generalized nuclear norm LRR}
Assume $X\not=0$ and $A=XZ$ have feasible solution(s), i.e.,
$A\in\mbox{Range}(X)$. Then
\begin{equation}
\label{equation: solution to noiseless nuclear LRR} Z^*=X^\dag A
\end{equation}
is the unique minimizer to the generalized heuristic LRR problem:
\begin{equation}
\label{equation: noiseless generalized nuclear LRR} \min_Z
||Z||_*,\ \ \mbox{s.t.}\ \ A=XZ.
\end{equation}
\end{lemma}

\begin{lemma}
\label{lemma: block-matrix} For any four matrices $B$, $C$, $D$
and $F$ of compatible dimensions, we have the inequalities
\begin{equation}
\begin{Vmatrix}
\begin{bmatrix}
B & C\\
D & F
\end{bmatrix}
\end{Vmatrix}_*
\ge||B||_*+||F||_*\mbox{ and } \begin{Vmatrix}
\begin{bmatrix}
B & C\\
D & F
\end{bmatrix}
\end{Vmatrix}_*
\ge||B||_*,
\end{equation}
where the second equality holds if and only if $C=0$, $D=0$, and
$F=0$.
\end{lemma}

Then we prove the following lemma.

\begin{lemma}
\label{lemma: nuclear norm and trace} For any square matrix
$Y\in\mathbb{R}^{n\times n}$, we have $||Y||_*\ge\mbox{trace}(Y)$,
where the equality holds if and only if $Y$ is positive
semi-definite.
\end{lemma}
\begin{proof} We prove by mathematical induction. When $n=1$, the
conclusion is clearly true. When $n=2$, we may simply write down
the singular values of $Y$ to prove.

Now suppose for any square matrix $\widetilde{Y}$, whose size does not
exceed $n-1$, the inequality holds. Then for any matrix
$Y\in\mathbb{R}^{n\times n}$, using Lemma \ref{lemma:
block-matrix}, we get
\begin{equation}
\begin{split}
||Y||_*&=
\begin{Vmatrix}
\begin{bmatrix}
Y_{11} & Y_{12}\\
Y_{21} & Y_{22}
\end{bmatrix}
\end{Vmatrix}_*\\
&\ge||Y_{11}||_*+||Y_{22}||_*\\
&\ge\mbox{trace}(Y_{11})+\mbox{trace}(Y_{22})\\
&=\mbox{trace}(Y),
\end{split}
\end{equation}
where the second inequality holds due to the inductive assumption
on the matrices $Y_{11}$ and $Y_{22}$. So we always have
$||Y||_*\ge\mbox{trace}(Y)$.

It is easy to check that any positive semi-definite matrix $Y$, it
satisfies $||Y||_*=\mbox{trace}(Y)$. On the other hand, just
following the above proof by choosing $Y_{22}$ as $2\times 2$
submatrices, we can easily get that $||Y||_*>\mbox{trace}(Y)$
strictly holds if $Y\in\mathbb{R}^{n\times n}$ is asymmetric. So
if $||Y||_*=\mbox{trace}(Y)$, then $Y$ must be symmetric. Then the
singular values of $Y$ are simply the absolute values of its
eigenvalues. As $\mbox{trace}(Y)$ equals the sum of all
eigenvalues of $Y$, $||Y||_*=\mbox{trace}(Y)$ holds only if all
the eigenvalues of $Y$ are non-negative. $\Box$
\end{proof}

Using Lemma~\ref{lemma: unique solution to generalized nuclear
norm LRR}, we may consider the following unconstrained problem
\begin{equation}
\label{equation: unconstrained problem} \min_{Z} f(Z)\triangleq
||Z||_*+||X(I-Z)X^\dag||_*,
\end{equation}
which is transformed from \eqref{equation: noiseless nuclear
LatLRR} be eliminating $L$ therein. Then we have the following
result.

\begin{proposition}
\label{proposition: minimal value} Unconstrained optimization
problem \eqref{equation: unconstrained problem} has a minimum
objective function value $\mbox{rank}(X)$.
\end{proposition}
\begin{proof}
Recall that the sub-differential of the nuclear norm of a matrix
$Z$ is~\cite{Cai-2008-SVT}
\begin{equation}
\label{equation: sub-differential of the nuclear norm}
\partial_{Z}||Z||_*=\{U_ZV_Z^T+R|U_Z^TR=0,RV_Z=0,||R||_2\le1\},
\end{equation}
where $U_Z\Sigma_Z V_Z^T$ is the skinny SVD of the matrix $Z$. We
prove that $Z^*=1/2X^\dag X$ is an optimal solution to
\eqref{equation: unconstrained problem}. It is sufficient to show
that
\begin{equation}
\begin{split}
0\in
\partial_Z
f(Z^*)&=\partial_Z||Z^*||_*+\partial_Z||X(I-Z^*)X^\dag||_*\\
&=\partial_Z||Z^*||_*-X^T\partial_{X(I-Z)X^\dag}||X(I-Z^*)X^\dag||_*(X^\dag)^T.
\end{split}
\end{equation}
Notice that $X(I-Z^*)X^\dag=U_X(1/2I)U_X^T$ is the skinny SVD of
$X(I-Z^*)X^\dag$ and $Z^*=V_X(1/2I)V_X^T$ is the skinny SVD of
$Z^*$. So $\partial_{Z}f(Z^*)$ contains
\begin{equation}
V_XV_X^T-X^T(U_XU_X^T)(X^\dag)^T =V_XV_X^T-V_X\Sigma_X
U_X^TU_XU_X^TU_X\Sigma_X^{-1}V_X^T=0.
\end{equation}

Substituting $Z^*=1/2X^\dag X$ into \eqref{equation: unconstrained
problem}, we get the minimum objective function value
$\mbox{rank}(X)$. $\Box$
\end{proof}

Next, we have the form of the optimal solutions to
\eqref{equation: unconstrained problem} as follows.

\begin{proposition}
\label{proposition: necessary condition of the solution} The
optimal solutions to the unconstrained optimization problem
\eqref{equation: unconstrained problem} can be written as
$Z^*=V_X\widehat{W}V_X^T$.
\end{proposition}
\begin{proof}
Let $(V_X)_\perp$ be the orthogonal complement of $V_X$. According
to Proposition \ref{proposition: minimal value}, $\mbox{rank}(X)$
is the minimum objective function value of \eqref{equation:
unconstrained problem}. Thus we get
\begin{equation}
\label{inequality: proof of necessary condition}
\begin{split}
\mbox{rank}(X)&=||Z^*||_*+||X(I-Z^*)X^\dag||_*\\
&=
\begin{Vmatrix}
\begin{bmatrix}
V_X^T\\
(V_X)_\perp^T
\end{bmatrix}
Z^*
\begin{bmatrix}
V_X,(V_X)_\perp
\end{bmatrix}
\end{Vmatrix}_*
+||X(I-Z^*)X^\dag||_*\\
&=
\begin{Vmatrix}
\begin{bmatrix}
V_X^TZ^*V_X & V_X^TZ^*(V_X)_\perp\\
(V_X)_\perp^TZ^*V_X & (V_X)_\perp^TZ^*(V_X)_\perp
\end{bmatrix}
\end{Vmatrix}_*
+||X(I-Z^*)X^\dag||_*\\
&\ge ||V_X^TZ^*V_X||_*+||U_X\Sigma_X V_X^T(I-Z^*)V_X\Sigma_X^{-1}U_X^T||_*\\
&=||V_XV_X^TZ^*V_XV_X^T||_*+||U_X\Sigma_X V_X^T(I-V_XV_X^TZ^*V_XV_X^T)V_X\Sigma_X^{-1}U_X^T||_*\\
&=||V_XV_X^TZ^*V_XV_X^T||_*+||X(I-V_XV_X^TZ^*V_XV_X^T)X^\dag||_*\\
&\ge\mbox{rank}(X),
\end{split}
\end{equation}
where the second inequality holds by viewing
$Z=V_XV_X^TZ^*V_XV_X^T$ as a feasible solution to \eqref{equation:
unconstrained problem}. Then all the inequalities in
\eqref{inequality: proof of necessary condition} must be
equalities. By Lemma \ref{lemma: block-matrix} we have
\begin{equation}
V_X^TZ^*(V_X)_\perp=(V_X)_\perp^TZ^*V_X=(V_X)_\perp^TZ^*(V_X)_\perp=0.
\end{equation}
That is to say
\begin{equation}
\begin{bmatrix}
V_X^T\\
(V_X)_\perp^T
\end{bmatrix}
Z^*
\begin{bmatrix}
V_X,(V_X)_\perp
\end{bmatrix}
=
\begin{bmatrix}
\widehat{W} & 0\\
0 & 0
\end{bmatrix},
\end{equation}
where $\widehat{W}=V_X^TZ^*V_X$. Hence the equality
\begin{equation}
Z^*=
\begin{bmatrix}
V_X,(V_X)_\perp
\end{bmatrix}
\begin{bmatrix}
\widehat{W} & 0\\
0 & 0
\end{bmatrix}
\begin{bmatrix}
V_X^T\\
(V_X)_\perp^T
\end{bmatrix}
=V_X\widehat{W}V_X^T
\end{equation}
holds. $\Box$
\end{proof}

Based on all the above lemmas and propositions, the following
proposition gives the whole closed form solutions to the
unconstrained optimization problem \eqref{equation: unconstrained
problem}. So the solution to problem \eqref{equation:
unconstrained problem} is non-unique.
\begin{proposition}
\label{proposition: sufficient and necessary condition for
unconstrained problem} The solutions to the unconstrained
optimization problem \eqref{equation: unconstrained problem} are
$Z^*=V_X\widehat{W}V_X^T$, where $\widehat{W}$ satisfies: 1. it is block
diagonal and its blocks are compatible with
$\Sigma_X$\footnote{Please refer to Theorem~\ref{theorem:
sufficient and necessary condition for nuclear norm LatLRR} for
the meaning of ``compatible with $\Sigma_X$."}; 2. both $\widehat{W}$
and $I-\widehat{W}$ are positive semi-definite.
\end{proposition}
\begin{proof}
First, we prove the sufficiency. Suppose $Z^*=V_X\widehat{W}V_X^T$
satisfies all the conditions in the theorem. Substitute it into
the objective function, we have
\begin{equation}
\begin{split}
||Z^*||_*+||X(I-Z^*)X^\dag||_*&=||\widehat{W}||_*+||\Sigma_X(I-\widehat{W})\Sigma_X^{-1}||_*\\
&=||\widehat{W}||_*+\mbox{trace}(\Sigma_X(I-\widehat{W})\Sigma_X^{-1})\\
&=||\widehat{W}||_*+\mbox{trace}(I-\widehat{W})\\
&=||\widehat{W}||_*+\mbox{rank}(X)-\mbox{trace}(\widehat{W})\\
&=\mbox{rank}(X)\\
&=\min_Z ||Z||_*+||X(I-Z)X^\dag||_*,
\end{split}
\end{equation}
where based on Lemma \ref{lemma: nuclear norm and trace} the
second and the fifth equalities hold since
$I-\widehat{W}=\Sigma_X(I-\widehat{W})\Sigma_X^{-1}$ as $\widehat{W}$ is block
diagonal and both $I-\widehat{W}$ and $\widehat{W}$ are positive
semi-definite.

Next, we give the proof of the necessity. Let $Z^*$ represent a
minimizer. According to Proposition \ref{proposition: necessary
condition of the solution}, $Z^*$ could be written as
$Z^*=V_X\widehat{W}V_X^T$. We will show that $\widehat{W}$ satisfies the
stated conditions. Based on Lemma \ref{lemma: nuclear norm and
trace}, we have
\begin{equation}
\begin{split}
\mbox{rank}(X)&=||Z^*||_*+||X(I-Z^*)X^\dag||_*\\
&=||\widehat{W}||_*+||\Sigma_X(I-\widehat{W})\Sigma_X^{-1}||_*\\
&\ge||\widehat{W}||_*+\mbox{trace}(\Sigma_X(I-\widehat{W})\Sigma_X^{-1})\\
&=||\widehat{W}||_*+\mbox{trace}(I-\widehat{W})\\
&=||\widehat{W}||_*+\mbox{rank}(X)-\mbox{trace}(\widehat{W})\\
&\ge\mbox{rank}(X).
\end{split}
\end{equation}
Thus all the inequalities above must be equalities. From the last
equality and Lemma~\ref{lemma: nuclear norm and trace}, we
directly get that $\widehat{W}$ is positive semi-definite. By the
first inequality and Lemma~\ref{lemma: nuclear norm and trace}, we
know that $\Sigma_X(I-\widehat{W})\Sigma_X^{-1}$ is symmetric, i.e.,
\begin{equation}
\frac{\sigma_i}{\sigma_j}[I-\widehat{W}]_{ij}=\frac{\sigma_j}{\sigma_i}[I-\widehat{W}]_{ij},
\end{equation}
where $\sigma_i$ represents the $i$th entry on the diagonal of
$\Sigma_X$. Thus if $\sigma_i\neq \sigma_j$, then
$[I-\widehat{W}]_{ij}=0$, i.e., $\widehat{W}$ is block diagonal and its
blocks are compatible with $\Sigma_X$. Notice that
$I-\widehat{W}=\Sigma_X(I-\widehat{W})\Sigma_X^{-1}$. By Lemma \ref{lemma:
nuclear norm and trace}, we get that $I-\widehat{W}$ is also positive
semi-definite. Hence the proof is completed. $\Box$
\end{proof}

Now we can prove Theorem \ref{theorem: sufficient and necessary
condition for nuclear norm LatLRR}.
\begin{proof}
Let $\widehat{W}$ satisfy all the conditions in the theorem. According
to Proposition \ref{proposition: necessary condition of the
solution}, since the row space of $Z^*=V_X\widehat{W}V_X^T$ belongs to
that of $X$, it is obvious that $(Z^*,X(I-Z^*)X^\dag)$ is feasible
to problem \eqref{equation: noiseless nuclear LatLRR}. Now suppose
that \eqref{equation: noiseless nuclear LatLRR} has a better
solution $(\widetilde{Z},\widetilde{L})$ than $(Z^*,L^*)$, i.e.,
\begin{equation}
X=X\widetilde{Z}+\widetilde{L}X,
\end{equation}
and
\begin{equation}
||\widetilde{Z}||_*+||\widetilde{L}||_*<||Z^*||_*+||L^*||_*.
\end{equation}
Fixing $Z$ in \eqref{equation: noiseless nuclear LatLRR} and by
Lemma \ref{lemma: unique solution to generalized nuclear norm
LRR}, we have
\begin{equation}
||\widetilde{Z}||_*+||(X-X\widetilde{Z})X^\dag||_*\le||\widetilde{Z}||_*+||\widetilde{L}||.
\end{equation}
Thus
\begin{equation}
||\widetilde{Z}||_*+||(X-X\widetilde{Z})X^\dag||_*<||Z^*||_*+||X(I-Z^*)X^\dag||_*.
\end{equation}
So we obtain a contradiction with respect to the optimality of
$Z^*$ in Proposition \ref{proposition: sufficient and necessary
condition for unconstrained problem}, hence proving the theorem.
$\Box$
\end{proof}

\subsection{Proof of Proposition~\ref{proposition:vality}}

\begin{proof}
Suppose the optimal formulation $Z^*=V_X\widehat{W}V_X^T$ in Theorem
\ref{theorem: sufficient and necessary condition for nuclear norm
LatLRR} could be written as
$Z^*=V_X\tilde{W}V_X^T+S_1\tilde{W}V_X^T$, where $\tilde{W}$ is
idempotent and $S_1$ satisfies $\tilde{W}V_X^TS_1=0$. Then we have
\begin{equation}
V_X\widehat{W}X_X^T=V_X\tilde{W}V_X^T+S_1\tilde{W}V_X^T.
\end{equation}
By multiplying both sides with $V_X^T$ and $V_X$ on the left and
right, respectively, we get
\begin{equation}
\widehat{W}=\tilde{W}+V_X^TS_1\tilde{W}.
\end{equation}
As a result, $\widehat{W}$ is idempotent:
\begin{equation}
\begin{split}
\widehat{W}^2&=(\tilde{W}+V_X^TS_1\tilde{W})(\tilde{W}+V_X^TS_1\tilde{W})\\
&=\tilde{W}^2+V_X^TS_1\tilde{W}^2+\tilde{W}V_X^TS_1\tilde{W}+V_X^TS_1\tilde{W}V_X^TS_1\tilde{W}\\
&=\tilde{W}+V_X^TS_1\tilde{W}+\tilde{W}V_X^TS_1\tilde{W}+V_X^TS_1\tilde{W}V_X^TS_1\tilde{W}\\
&=\tilde{W}+V_X^TS_1\tilde{W}=\widehat{W},
\end{split}
\end{equation}
which is contradictory to the assumption. $\Box$
\end{proof}

\section{Conclusions}
\label{section: Conclusion}

Based on the expositions in Section \ref{section: Analysis on Rank
Minimization Model} and the proofs in Section
\ref{section:proofs}, we conclude that even for rank minimization problems as simple as
noiseless LatLRR, replacing rank with nuclear norm is not valid.
We have also found that LatLRR is actually problematic
because the solution to its nuclear norm minimization formation is not unique. We can
also have the following interesting connections between LRR and
LatLRR. Namely, LatLRR is indeed an extension of LRR because its
solution set strictly includes that of LRR, no matter for the rank minimization problem
or the nuclear norm minimization formulation. So we can summarize their relationship as
Figure~\ref{figure: corresponding relationships between different
LRR and LatLRR model}.
\begin{figure}
\centering
\includegraphics[width=0.7\textwidth]{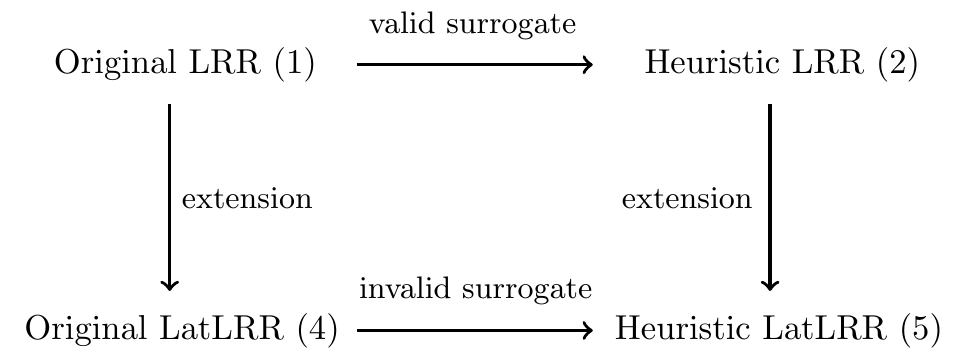}
\caption{The detailed relationship among the original LRR
\eqref{equation: noiseless rank LRR}, the heuristic LRR
\eqref{equation: noiseless nuclear LRR}, the original LatLRR
\eqref{equation: noiseless original LatLRR}, and the heuristic
LatLRR \eqref{equation: noiseless nuclear LatLRR} in the sense of
their solution sets.} \label{figure: corresponding relationships
between different LRR and LatLRR model}
\end{figure}

Although the existing formulation of LatLRR is imperfect, since
some scholars have demonstrated its effectiveness in subspace
clustering by using a solution which is randomly chosen in some
sense, in the future we will consider how to choose the best
solution in the solution set in order to further improve the
performance of LatLRR.

\comment{

Based on the proofs in Section \ref{section: Analysis on Rank
Minimization Model} and Section \ref{section: Analysis on Nuclear
Norm Minimization Model}, we conclude the following relationships
as Fig. \ref{figure: corresponding relationships between different
LRR and LatLRR model}, which provide us with an impressive
conclusion: For LatLRR, the nuclear norm problem fails to well
approach the rank one, because of its non-unique minimizers and
inconsistent solutions with the rank problem. By taking into
consideration that the solutions to LatLRR strictly include those
of LRR, we conclude that LatLRR is actually an extending model
both on the rank function and on the nuclear norm.

To sum up, in this paper, by theoretically analyzing the complete
solutions to LatLRR model both on the rank function and the
nuclear norm, we take a counterexample to show that nuclear norm
is not always a good surrogate for the rank. By adding more
constraints, future works could be expected on picking up
appropriate solution for LatLRR, such as combining with the
technique of semi-supervised learning.

}

\subsubsection*{Acknowledgments.}
Hongyang Zhang and Chao Zhang are supported by National Key Basic Research Project of China (973 Program) 2011CB302400 and National Nature Science Foundation of China (NSFC Grant, no. 61071156). Zhouchen Lin is supported by National Nature Science Foundation of China (Grant nos. 61272341, 61231002, and 61121002).

\bibliography{reference}
\bibliographystyle{splncs}
\end{document}